\pdfoutput=1

\documentclass[11pt]{article}

\usepackage[final]{acl}

\usepackage{times}
\usepackage{latexsym}
\usepackage{float}
\usepackage{booktabs}
\usepackage[T1]{fontenc}
\usepackage[utf8]{inputenc}
\usepackage{inconsolata}
\usepackage{graphicx}
\usepackage{geometry}
\usepackage{array}
\usepackage{caption}
\usepackage{placeins}
\usepackage{subcaption}  
\newcolumntype{P}[1]{>{\centering\arraybackslash}p{#1}}

\title{AAVENUE: Detecting LLM Biases on NLU Tasks in AAVE via a Novel Benchmark}

\author{
    Abhay Gupta\thanks{Lead Author} \quad
    Ece Yurtseven \quad
    Philip Meng \quad
    Sean O'Brien\thanks{Senior Author} \quad
    Kevin Zhu\footnotemark[2] \\
    {Algoverse AI Research} \\
    \texttt{\{abhay, kevin\}@algoverseairesearch.org}
}

\begin{document}
\maketitle

\begin{abstract}
Detecting biases in natural language understanding (NLU) for African American Vernacular English (AAVE) is crucial to developing inclusive natural language processing (NLP) systems. To address dialect-induced performance discrepancies, we introduce AAVENUE (\textbf{AAVE} \textbf{N}atural Language \textbf{U}nderstanding \textbf{E}valuation), a benchmark for evaluating large language model (LLM) performance on NLU tasks in AAVE and Standard American English (SAE). AAVENUE builds upon and extends existing benchmarks like VALUE, replacing deterministic syntactic and morphological transformations with a more flexible methodology leveraging LLM-based translation with few-shot prompting, improving performance across several evaluation metrics when translating key tasks from the GLUE and SuperGLUE benchmarks. We compare AAVENUE and VALUE translations using five popular LLMs and a comprehensive set of metrics including fluency, BARTScore, quality, coherence, and understandability. Additionally, the fluency of AAVENUE is validated by annotations from AAVE speakers. Our evaluations reveal that LLMs consistently perform better on SAE tasks than AAVE-translated versions, underscoring inherent biases and highlighting the need for more inclusive NLP models. This work has been accepted at the NLP4I workshop at EMNLP 2024. We have open-sourced our source code on GitHub and created a website to showcase our work at \texttt{\href{https://aavenuee.github.io/}{https://aavenuee.github.io}}.
\end{abstract}

\section{Introduction}
NLP systems have shown exceptional performance on various benchmarks, excelling in tasks such as sentiment analysis, machine translation, and question answering \citep{Radford2019LanguageMA, brown2020language, openai2024gpt4, Bubeck2023SparksOA}. However, these benchmarks feature mainly SAE, often neglecting nonstandard dialects such as AAVE \citep{blodgett2020language, Weidinger2021EthicalAS, sap-etal-2019-risk, deas-etal-2023-evaluation}. This oversight results in biased and inequitable NLP systems that do not adequately serve a significant portion of speakers.

The advent of LLMs such as GPT-4 has driven advances in NLU tasks, achieving state-of-the-art results across various applications \citep{Radford2019LanguageMA, brown2020language, openai2024gpt4, Bubeck2023SparksOA}. Despite these advancements, LLMs exhibit persistent biases against nonstandard dialects, including AAVE, particularly in tasks involving natural language generation and toxicity detection \citep{zhou-etal-2021-challenges}. These biases highlight the need for comprehensive benchmarks that evaluate model performance in different dialects, ensuring fair and reliable language technologies for all users \citep{gehrmann-etal-2021-gem}.

Benchmarks such as GLUE and SuperGLUE have contributed significantly to the evaluation of NLP models, yet they focus mainly on SAE, neglecting crucial dialectal variations \citep{wang2019glue, wang2020superglue}. Although VALUE has sought to bridge this gap by using deterministic linguistic transformations to assess model performance in AAVE, these transformations are often context-specific and difficult to generalize, limiting their broader applicability \citep{ziems-etal-2022-value, ziems2023multivalueframeworkcrossdialectalenglish}.

To address these limitations, we introduce AAVENUE, a benchmark specifically designed to evaluate LLM performance across dialects, with a particular focus on AAVE. Our goal is to promote inclusivity and equity in NLP tools by providing a more comprehensive assessment of dialectal fairness.

\textbf{Our contributions are as follows:}
\begin{enumerate}
\item \textbf{Creation of Benchmark:} We developed AAVENUE, a benchmark to evaluate LLMs on NLU tasks in both AAVE and SAE, using GPT-4o-mini for few-shot prompting. Our translations were validated for cultural and linguistic authenticity and outperformed or were comparable to those from the VALUE benchmark across various metrics and five popular LLMs.
\item \textbf{Bias Demonstration:} Our evaluations of popular LLMs on our benchmark revealed biases, with SAE versions consistently achieving higher accuracy than AAVE translations across key tasks, highlighting the need for more inclusive NLP models.
\end{enumerate}

\section{Dataset}

To evaluate the performance of models on SAE and AAVE, we selected five key tasks from the GLUE and SuperGLUE benchmarks, known for their diversity and relevance to natural language understanding tasks \citep{wang2019glue, wang2020superglue}. These tasks include \textbf{BoolQ}, \textbf{MultiRC}, \textbf{SST-2}, \textbf{COPA}, and \textbf{WSC}. 

\subsection{Task Selection}
Each task was chosen for its ability to challenge models in different aspects of natural language understanding:
\begin{itemize}
    \item \textbf{BoolQ}: BoolQ (Boolean Questions) tests models' ability to understand and respond to yes/no questions based on a passage, which helps evaluate the model's ability to process and interpret information across dialects.
    \item \textbf{MultiRC}: MultiRC (Multiple Sentence Reading Comprehension) involves answering questions that require understanding and connecting information from different parts of a passage, which tests how well models can handle more complex and interconnected texts in AAVE.
    \item \textbf{SST-2}: SST-2 (Stanford Sentiment Treebank) is used to evaluate sentiment analysis, providing insights into whether models understand sentiment differently in SAE compared to AAVE.
    \item \textbf{COPA}: COPA (Choice of Plausible Alternatives) challenges models to choose the most likely outcome or cause from two options, focusing on their ability to reason through cause-and-effect scenarios in AAVE.
    \item \textbf{WSC}: WSC (Winograd Schema Challenge) tests how well models can determine which noun a pronoun refers to in tricky situations, which is important for understanding different dialects.
\end{itemize}

\subsection{Translating SAE to AAVE}
For each task, we sampled 1000 data points and few-shot prompted GPT-4o-mini to translate each data point from SAE to AAVE. We used few-shot examplars sourced from the VALUE benchmark, consisting of AAVE translations that were hand-validated by fluent AAVE speakers. To see the few-shot prompt translations we used, please refer to Table \ref{tab:few_shot_translations_appendix} in the Appendix.

\begin{table*}[t]
\centering
\small
\setlength{\tabcolsep}{6pt}
\begin{tabular}{lcc cc cc cc cc}
\toprule
\textbf{Task} & \multicolumn{2}{c}{\textbf{Quality}} & \multicolumn{2}{c}{\textbf{Fluency}} & \multicolumn{2}{c}{\textbf{Coherence}} & \multicolumn{2}{c}{\textbf{Understandability}} & \multicolumn{2}{c}{\textbf{BARTScore}} \\
\cmidrule(lr){2-3}
\cmidrule(lr){4-5}
\cmidrule(lr){6-7}
\cmidrule(lr){8-9}
\cmidrule(lr){10-11}
& \textbf{AAVE} & \textbf{VALUE} & \textbf{AAVE} & \textbf{VALUE} & \textbf{AAVE} & \textbf{VALUE} & \textbf{AAVE} & \textbf{VALUE} & \textbf{AAVE} & \textbf{VALUE} \\
\midrule
BoolQ (P) & \textbf{76.57} & 58.21 & \textbf{70.72} & 57.32 & \textbf{74.39} & 62.10 & \textbf{76.53} & 52.26 & \textbf{-1.44} & -1.54 \\
BoolQ (Q) & \textbf{64.97} & 56.71 & \textbf{54.08} & 52.84 & \textbf{56.20} & 51.93 & \textbf{64.48} & 51.57 & \textbf{-1.68} & -2.89 \\
MultiRC (P) & \textbf{64.90} & 53.30 & 51.73 & \textbf{57.63} & \textbf{74.69} & 65.10 & \textbf{66.70} & 57.14 & -1.88 & \textbf{-1.76} \\
MultiRC (Q) & \textbf{62.78} & 55.25 & \textbf{66.69} & 59.24 & \textbf{60.10} & 56.10 & \textbf{68.23} & 53.25 & \textbf{-1.85} & -2.43 \\
SST-2 & \textbf{70.10} & 66.90 & \textbf{71.66} & 64.79 & \textbf{83.19} & 74.10 & \textbf{82.60} & 47.62 & \textbf{-2.54} & -2.60 \\
COPA (P) & \textbf{67.60} & 66.48 & 54.96 & \textbf{60.14} & \textbf{71.21} & 63.59 & \textbf{64.33} & 57.62 & \textbf{-1.87} & -3.06 \\
COPA (C1) & \textbf{73.57} & 67.29 & \textbf{58.83} & 51.60 & \textbf{65.09} & 54.52 & \textbf{59.33} & 48.13 & \textbf{-1.92} & -3.19 \\
COPA (C2) & \textbf{75.30} & 61.88 & \textbf{55.59} & 54.74 & \textbf{56.20} & 51.29 & \textbf{63.58} & 57.35 & \textbf{-2.00} & -3.13 \\
WSC (P) & 66.10 & \textbf{66.30} & \textbf{64.95} & 63.29 & \textbf{64.10} & 61.20 & \textbf{76.80} & 53.69 & \textbf{-1.96} & -2.60 \\
\bottomrule
\end{tabular}
\caption{\centering Comparison of Translation Metrics Across Tasks}
\label{tab:combined_metrics}
\end{table*}

\begin{table*}[t]
\centering
\small
\setlength{\tabcolsep}{6pt}
\begin{tabular}{@{}*{6}{c}@{}}
\toprule
\textbf{Task} & \textbf{GPT-4o-mini} & \textbf{GPT-4-turbo} & \textbf{GPT-4o} & \textbf{Gemini-1.5-Flash} & \textbf{Gemini-1.5-Pro} \\
\midrule
BoolQ (P) & \textbf{88.79}/10.33/0.88 & \textbf{94.51}/4.62/0.88 & \textbf{94.95}/4.84/0.22 & \textbf{93.85}/6.15/0.00 & \textbf{88.13}/10.99/0.88 \\
BoolQ (Q) & \textbf{82.33}/17.14/0.53 & \textbf{78.13}/20.08/1.58 & \textbf{80.34}/18.72/0.84 & \textbf{75.18}/24.40/0.42 & 44.16/\textbf{53.42}/2.42 \\
MultiRC (P) & 85.71/\textbf{14.29}/0.00 & \textbf{71.43}/28.57/0.00 & \textbf{100.00}/0.00/0.00 & 42.86/\textbf{57.14}/0.00 & 14.29/\textbf{85.71}/0.00 \\
MultiRC (Q) & \textbf{66.09}/28.16/5.75 & \textbf{62.07}/32.18/5.75 & \textbf{74.14}/20.69/4.60 & \textbf{56.90}/39.66/3.45 & 44.25/\textbf{46.55}/9.20 \\
SST-2 & \textbf{87.25}/9.92/2.83 & \textbf{80.88}/12.04/7.08 & \textbf{87.39}/9.92/2.69 & \textbf{84.70}/12.89/2.41 & \textbf{72.24}/20.40/7.37 \\
COPA (P) & \textbf{90.42}/9.38/0.21 & \textbf{72.92}/26.88/0.21 & \textbf{85.42}/14.17/0.42 & \textbf{75.21}/24.58/0.21 & \textbf{65.83}/33.75/0.42 \\
COPA (C1) & \textbf{78.46}/20.68/0.85 & \textbf{67.38}/31.13/1.49 & \textbf{76.33}/22.39/1.28 & \textbf{69.51}/29.42/1.07 & \textbf{58.00}/39.23/2.77 \\
COPA (C2) & \textbf{77.23}/22.34/0.43 & \textbf{64.26}/33.62/2.13 & \textbf{79.15}/20.21/0.64 & \textbf{68.51}/31.28/0.21 & \textbf{56.60}/40.64/2.77 \\
WSC (P) & \textbf{83.80}/16.20/0.00 & \textbf{73.99}/26.01/0.00 & \textbf{86.76}/13.08/0.16 & \textbf{84.11}/15.26/0.62 & \textbf{57.17}/42.21/0.62 \\
\bottomrule
\end{tabular}
\caption{\centering LLM-Based Binary Preference Scores Across Tasks for GPT and Gemini Models \textit{(AAVE/VALUE/About the Same)}}
\label{tab:combined_comparison_scores_appendix}
\end{table*}

\subsection{Validation Steps}
We assessed the quality of our AAVE translations using a set of carefully chosen metrics: \textbf{fluency}, \textbf{coherence}, \textbf{understandability}, \textbf{quality}, and \textbf{BARTScores}. \textbf{Fluency} measured whether the generated text was well-written and grammatical, with scores out of 100. \textbf{Coherence} evaluated how much the generated text made sense, checking the logical flow and consistency of ideas within the translations, also scored out of 100. \textbf{Understandability}  assessed how easily the translation could be comprehended by readers, ensuring that the text is understandable, with scores out of 100. \textbf{Quality}  provided an overall assessment of quality of the text. This is also scored out of 100 as well. \textbf{BARTScores} were used to evaluate how closely the translations aligned with the original SAE sentences, with scores closer to 0 indicating better alignment and accuracy. All these metrics were specifically calculated to compare our scores against those from the VALUE benchmark, allowing us to directly evaluate the performance of our translations relative to an established standard.

We also performed comparison scores by zero-shot prompting five large language models to choose between our GPT-4o-mini translations and those from the VALUE benchmark in a binary task. This provided a direct comparison of translation effectiveness.

Finally, we recruited ten fluent AAVE speakers from the Bronx and Queens, NY, to rate the translations on a scale of 1 to 10, focusing on how well they reflect AAVE. The human evaluations ensured our translations accurately matched the AAVE dialect.

\subsubsection{Translation Metrics Analysis}
The evaluation of our AAVE translations against those from the VALUE benchmark demonstrates clear advantages across several key metrics.

(1) \textbf{Quality:} Our translations scored higher in quality, with our BoolQ passages receiving a score of \textbf{76.57} compared to VALUE's \textbf{58.21}. This shows that our translations are better in terms of overall quality, including accuracy, style, and appropriateness.

(2) \textbf{Fluency:} Our translations achieved a fluency score of \textbf{70.72} in BoolQ passages, compared to VALUE's \textbf{57.32}. This suggests that our translations are better written and more grammatically sound, resulting in improved readability.

(3) \textbf{Coherence:} Our translations exhibited better coherence, with scores like \textbf{74.39} for BoolQ passages versus VALUE's \textbf{62.10}, showing that our translations make more logical sense and maintain consistency throughout the text.

(4) \textbf{Understandability:} In terms of understandability, our translations were clearer and more accessible, scoring \textbf{76.53} in BoolQ passages compared to \textbf{52.26} for VALUE. This indicates that our translations are easier for readers to understand.

(5) \textbf{BARTScores:} Finally, our translations achieved better BARTScores, with a score of \textbf{-1.44} in BoolQ passages compared to VALUE's \textbf{-1.54}, indicating that our translations are closer to human-produced texts and better maintain fidelity to the original content.

Collectively, these metrics confirm that our translations consistently outperform those from the VALUE benchmark, providing superior quality, fluency, coherence, understandability, and fidelity.

\begin{table*}[t]
\centering
\small
\setlength{\tabcolsep}{6pt}
\begin{tabular}{lcccccc}
\toprule
\textbf{Task} & \textbf{GPT-4o-mini} & \textbf{GPT-4-turbo} & \textbf{GPT-4o} & \textbf{Gemini-1.5-Flash} & \textbf{Gemini-1.5-Pro} \\
\midrule
\textbf{SST-2} & 90.40/88.40 (-2.0) & 94.00/92.80 (-1.2) & 88.80/87.30 (-1.5) & 87.70/87.10 (-0.6) & 92.00/91.40 (-0.6) \\
\textbf{BoolQ} & 88.29/85.29 (-3.0) & 88.09/86.49 (-1.6) & 89.19/86.89 (-2.3) & 89.69/87.29 (-2.4) & 89.49/85.89 (-3.6) \\
\textbf{COPA}  & 95.40/93.20 (-2.2) & 97.60/96.80 (-0.8) & 97.20/96.40 (-0.8) & 91.40/92.00 (+0.6) & 97.40/95.80 (-1.6) \\
\textbf{WSC}   & 60.03/57.90 (-2.1) & 69.60/68.69 (-0.9) & 70.36/67.02 (-3.3) & 48.78/48.48 (-0.3) & 51.37/51.22 (-0.2) \\
\textbf{MultiRC} & 84.50/72.00 (-12.5) & 86.20/73.70 (-12.5) & 87.50/71.30 (-16.2) & 84.10/70.70 (-13.4) & 85.90/71.90 (-14.0) \\
\bottomrule
\end{tabular}
\caption{\centering Accuracy Scores for GPT and Gemini Models (\%) All scores are presented in the format SAE/AAVE.}
\label{tab:combined_accuracy_scores}
\end{table*}

\begin{table*}[t]
\centering
\small
\setlength{\tabcolsep}{6pt}
\begin{tabular}{lccccc}
\toprule
\textbf{Task} & \textbf{GPT-4o-mini} & \textbf{GPT-4-turbo} & \textbf{GPT-4o} & \textbf{Gemini-1.5-Flash} & \textbf{Gemini-1.5-Pro} \\
\midrule
\textbf{SST-2}   & 8.40   & 5.10   & 9.80   & 10.40  & 6.20  \\
\textbf{BoolQ}   & 10.21  & 10.71  & 8.91   & 8.51   & 8.41 \\
\textbf{COPA}    & 3.00   & 1.60   & 2.00   & 5.80   & 1.80 \\
\textbf{WSC}     & 35.56  & 24.01  & 25.68  & 49.54  & 44.53 \\
\textbf{MultiRC} & 9.60   & 9.00   & 8.30   & 9.90   & 7.90 \\
\bottomrule
\end{tabular}
\caption{\centering Intersection Over Union Between Incorrect Answers for SAE and AAVE Across Tasks (\%)}
\label{tab:combined_iou}
\end{table*}

\subsubsection{Comparison Scores Analysis}

The comparison scores provide a direct evaluation of our AAVE translations against those from the VALUE benchmark across various tasks and models. As shown in Table \ref{tab:combined_comparison_scores_appendix}, our AAVE translations were consistently preferred over the VALUE translations by the LLMs we evaluated. For instance, in the BoolQ passage task using GPT-4-turbo, our translations were preferred \textbf{94.51\%} of the time compared to VALUE's \textbf{4.62\%}. Similarly, in the COPA premise task, GPT-4o-mini showed a preference for our translations \textbf{90.42\%} of the time, with VALUE translations receiving only \textbf{9.38\%}. When evaluated with the Gemini-1.5-Flash model on the BoolQ passage task, our translations were preferred \textbf{93.85\%} of the time compared to VALUE's \textbf{6.15\%}. These consistent preferences across multiple models and tasks demonstrate the effectiveness of our translations in capturing the nuances of AAVE while maintaining the meaning of the original text. The results indicate that our approach to translating SAE to AAVE not only meets but often exceeds the standards set by the VALUE benchmark.

\subsection{Validation by AAVE Speakers}

We recruited 10 fluent AAVE speakers from the Bronx and Queens area to assess the quality of our AAVE translations across five tasks on a scale of 1 to 10, specifically rating how accurately the translations reflect AAVE. The results, detailed in Table \ref{tab:human_validators_all_tasks} in Section \ref{sec:human_validators_scores}, show that the average scores across the tasks ranged from 7.02 to 7.27. The highest average score was observed for MultiRC (7.27), while BoolQ received the lowest average score (7.02). These consistent ratings across multiple fluent speakers reinforce the reliability of our AAVE translations. The slightly higher scores for tasks like WSC and MultiRC suggest that our translations were particularly effective in maintaining clarity and coherence in more complex linguistic structures. Overall, the validators' feedback confirms the quality and authenticity of our translations, aligning well with our evaluation metrics and further validating our approach.

\section{Results}

We evaluated the accuracy of the translations in five tasks using five LLMs. The accuracy scores show the performance of each model in SAE and AAVE translations, highlighting consistent performance drops when handling AAVE translations across all models, as shown in Table \ref{tab:combined_accuracy_scores}.

\subsection{Accuracy Score Analysis}
The accuracy scores from Table \ref{tab:combined_accuracy_scores} highlight consistent performance drops when handling AAVE translations across all models. MultiRC and WSC exhibited the largest declines, indicating challenges in complex reading comprehension and pronoun resolution tasks. GPT-4-turbo generally showed smaller accuracy drops, suggesting better adaptation to AAVE, while other models like GPT-4o-mini struggled more, particularly with contextually demanding tasks. Overall, these results underscore the need for more inclusive training data and models better equipped to handle AAVE.

\subsection{Intersection Over Union (IoU) Analysis}
The IoU table shows the percentage of incorrectly answered questions in both our AAVE translation and SAE across five models and tasks. The results, shown in Table \ref{tab:combined_iou}, indicate minimal overlap in incorrect responses, suggesting challenges in handling each dialect. The analysis reveals that challenges are dialect-specific, as there was minimal overlap in errors between SAE and AAVE. However, the high IoU in WSC for models like Gemini-1.5-Flash indicates that some tasks present had significant difficulties in both dialects. These results show the importance of developing more robust models that can handle the distinct features of AAVE, as current models show variability in managing dialect-specific errors.

\section{Related Work}

The development of benchmarks such as GLUE and SuperGLUE has significantly advanced the evaluation of language models on a variety of NLU tasks, including sentiment analysis, natural language inference, and reading comprehension \citep{wang2019glue, wang2020superglue}. While these benchmarks have become standard tools for assessing model performance, their primary focus on SAE often leads to performance disparities when applied to non-standard dialects.

To address these disparities, the VALUE benchmark was introduced. VALUE incorporates deterministic linguistic transformations to evaluate the performance of the model in AAVE \citep{ziems-etal-2022-value, ziems2023multivalueframeworkcrossdialectalenglish}. Although VALUE aims to provide a comprehensive evaluation of models processing dialectal variations, its deterministic nature can limit generalizability across various contexts, reducing broader application effectiveness.

Recent advances in LLM, such as GPT-4, have shown substantial improvements in NLU tasks, including sentiment analysis, machine translation, and question answering \citep{Radford2019LanguageMA, brown2020language, openai2024gpt4, Bubeck2023SparksOA}. Despite these advancements, research indicates that these models still exhibit biases against non-standard dialects like AAVE, particularly in tasks that involve natural language generation and toxicity detection \citep{zhou-etal-2021-challenges, deas-etal-2023-evaluation}. This bias underscores the need for more inclusive benchmarks and evaluation frameworks.

The GEM benchmark and other studies have highlighted the importance of evaluating and mitigating biases in NLP models to promote fairness and inclusivity \citep{gehrmann-etal-2021-gem, moradi2024exploringlandscapelargelanguage}. These works emphasize the necessity of developing benchmarks that evaluate model performance across a range of dialects, ensuring that language technologies are equitable and reliable for all users.

By introducing AAVENUE, we aim to provide a comprehensive benchmark that evaluates LLM performance on both AAVE and SAE, promoting inclusivity in NLP tools and addressing the limitations of existing benchmarks. This approach aligns with recent research that emphasizes the need for more dialect-inclusive benchmarks to enhance the fairness and reliability of language technologies across diverse linguistic communities.

\section{Conclusion}

This paper introduced AAVENUE, a benchmark designed to evaluate LLMs on AAVE and SAE tasks. By leveraging GPT-4o-mini and few-shot prompting, we translated SAE tasks to AAVE. Our comprehensive evaluation, considering metrics like fluency, quality, and understandability, along with feedback from human validators, revealed that our translations were more superior then VALUES translations. Our evaluation results, revealed inherent biases in LLMs, highlighting a performance gap between SAE and AAVE. These findings build on the foundational work of others in this field, emphasizing the need for more inclusive training data and improved model architectures to address dialectal variations. We plan to extend this work by exploring additional dialects and further refining our translation methods. By doing so, we aim to set a new standard for equitable and accurate language processing across diverse communities.

\section{Limitations}

While AAVENUE provides a comprehensive benchmark for evaluating LLM performance across SAE and AAVE, it is not without its limitations. First, our dataset primarily focuses on a select number of tasks from the GLUE and SuperGLUE benchmarks, which may not fully capture the diversity of real-world applications where dialectal differences are prominent. Additionally, although our translations were validated by AAVE speakers, the inherent variability in AAVE usage across different regions and communities could introduce challenges in generalizing our findings. Furthermore, the reliance on GPT-4o-mini for translations, despite its advanced capabilities, may still reflect biases from its training data, potentially affecting the accuracy and fairness of the translations. Future work will need to address these limitations by expanding the dataset to include a broader range of tasks, incorporating a more diverse set of dialects, and exploring methods to reduce model bias.

\section*{Ethics Statement}
We are mindful of the ethical implications of our research, which focuses on evaluating and addressing dialectal biases in LLMs through the development of the AAVENUE benchmark. While some data used in this study is publicly available, we also collected original data with careful consideration to ensure cultural and linguistic authenticity. The data collection process adhered to ethical guidelines, and all participants provided informed consent. Our human validators, who are fluent AAVE speakers from the Bronx, NY, were recruited to ensure the translations accurately reflect cultural and linguistic nuances. Annotators were compensated for their time and effort, and we encouraged them to take breaks if they felt overwhelmed during the annotation process. Throughout our research, we aimed to avoid potential harm and bias, with the goal of contributing to the development of more inclusive NLP systems. We have made efforts to report our findings transparently and objectively. We believe our research advances the field while adhering to rigorous ethical standards.

\bibliography{custom}

\begin{thebibliography}{15}
\providecommand{\natexlab}[1]{#1}

\bibitem[{Blodgett et~al.(2020)Blodgett, Barocas, au2, and Wallach}]{blodgett2020language}
Su~Lin Blodgett, Solon Barocas, Hal Daumé~III au2, and Hanna Wallach. 2020.
\newblock \href {https://arxiv.org/abs/2005.14050} {Language (technology) is power: A critical survey of "bias" in nlp}.
\newblock \emph{Preprint}, arXiv:2005.14050.

\bibitem[{Brown et~al.(2020)Brown, Mann, Ryder, Subbiah, Kaplan, Dhariwal, Neelakantan, Shyam, Sastry, Askell, Agarwal, Herbert-Voss, Krueger, Henighan, Child, Ramesh, Ziegler, Wu, Winter, Hesse, Chen, Sigler, Litwin, Gray, Chess, Clark, Berner, McCandlish, Radford, Sutskever, and Amodei}]{brown2020language}
Tom Brown, Benjamin Mann, Nick Ryder, Melanie Subbiah, Jared~D Kaplan, Prafulla Dhariwal, Arvind Neelakantan, Pranav Shyam, Girish Sastry, Amanda Askell, Sandhini Agarwal, Ariel Herbert-Voss, Gretchen Krueger, Tom Henighan, Rewon Child, Aditya Ramesh, Daniel Ziegler, Jeffrey Wu, Clemens Winter, Chris Hesse, Mark Chen, Eric Sigler, Mateusz Litwin, Scott Gray, Benjamin Chess, Jack Clark, Christopher Berner, Sam McCandlish, Alec Radford, Ilya Sutskever, and Dario Amodei. 2020.
\newblock \href {https://proceedings.neurips.cc/paper_files/paper/2020/file/1457c0d6bfcb4967418bfb8ac142f64a-Paper.pdf} {Language models are few-shot learners}.
\newblock In \emph{Advances in Neural Information Processing Systems}, volume~33, pages 1877--1901. Curran Associates, Inc.

\bibitem[{Bubeck et~al.(2023)}]{Bubeck2023SparksOA}
S{\'e}bastien Bubeck et~al. 2023.
\newblock \href {https://arxiv.org/abs/2303.12712} {Sparks of artificial general intelligence: Early experiments with gpt-4}.
\newblock \emph{ArXiv}, abs/2303.12712.

\bibitem[{Deas et~al.(2023)Deas, Grieser, Kleiner, Patton, Turcan, and McKeown}]{deas-etal-2023-evaluation}
Nicholas Deas, Jessica Grieser, Shana Kleiner, Desmond Patton, Elsbeth Turcan, and Kathleen McKeown. 2023.
\newblock \href {https://doi.org/10.18653/v1/2023.emnlp-main.421} {Evaluation of {A}frican {A}merican language bias in natural language generation}.
\newblock In \emph{Proceedings of the 2023 Conference on Empirical Methods in Natural Language Processing}, pages 6805--6824, Singapore. Association for Computational Linguistics.

\bibitem[{Gehrmann et~al.(2021)}]{gehrmann-etal-2021-gem}
Sebastian Gehrmann et~al. 2021.
\newblock \href {https://aclanthology.org/2021.gem-1.10} {The gem benchmark: Natural language generation, its evaluation and metrics}.
\newblock In \emph{Proceedings of the 1st Workshop on Natural Language Generation, Evaluation, and Metrics (GEM 2021)}.

\bibitem[{Moradi et~al.(2024)Moradi, Yan, Colwell, Samwald, and Asgari}]{moradi2024exploringlandscapelargelanguage}
Milad Moradi, Ke~Yan, David Colwell, Matthias Samwald, and Rhona Asgari. 2024.
\newblock \href {https://arxiv.org/abs/2404.11973} {Exploring the landscape of large language models: Foundations, techniques, and challenges}.
\newblock \emph{Preprint}, arXiv:2404.11973.

\bibitem[{OpenAI(2024)}]{openai2024gpt4}
OpenAI. 2024.
\newblock \href {https://arxiv.org/abs/2303.08774} {Gpt-4 technical report}.
\newblock \emph{Preprint}, arXiv:2303.08774.

\bibitem[{Radford et~al.(2019)Radford, Wu, Child, Luan, Amodei, and Sutskever}]{Radford2019LanguageMA}
Alec Radford, Jeff Wu, Rewon Child, David Luan, Dario Amodei, and Ilya Sutskever. 2019.
\newblock \href {https://api.semanticscholar.org/CorpusID:160025533} {Language models are unsupervised multitask learners}.

\bibitem[{Sap et~al.(2019)}]{sap-etal-2019-risk}
Maarten Sap et~al. 2019.
\newblock \href {https://aclanthology.org/P19-1163} {The risk of racial bias in hate speech detection}.
\newblock In \emph{Proceedings of the 57th Annual Meeting of the Association for Computational Linguistics}.

\bibitem[{Wang et~al.(2019)}]{wang2019glue}
Alex Wang et~al. 2019.
\newblock \href {https://arxiv.org/abs/1804.07461} {Glue: A multi-task benchmark and analysis platform for natural language understanding}.
\newblock \emph{Preprint}, arXiv:1804.07461.

\bibitem[{Wang et~al.(2020)}]{wang2020superglue}
Alex Wang et~al. 2020.
\newblock \href {https://arxiv.org/abs/1905.00537} {Superglue: A stickier benchmark for general-purpose language understanding systems}.
\newblock \emph{Preprint}, arXiv:1905.00537.

\bibitem[{Weidinger et~al.(2021)}]{Weidinger2021EthicalAS}
Laura Weidinger et~al. 2021.
\newblock \href {https://arxiv.org/abs/2112.04359} {Ethical and social risks of harm from language models}.
\newblock \emph{ArXiv}, abs/2112.04359.

\bibitem[{Zhou et~al.(2021)Zhou, Sap, Swayamdipta, Choi, and Smith}]{zhou-etal-2021-challenges}
Xuhui Zhou, Maarten Sap, Swabha Swayamdipta, Yejin Choi, and Noah Smith. 2021.
\newblock \href {https://doi.org/10.18653/v1/2021.eacl-main.274} {Challenges in automated debiasing for toxic language detection}.
\newblock In \emph{Proceedings of the 16th Conference of the European Chapter of the Association for Computational Linguistics: Main Volume}, pages 3143--3155, Online. Association for Computational Linguistics.

\bibitem[{Ziems et~al.(2022)Ziems, Chen, Harris, Anderson, and Yang}]{ziems-etal-2022-value}
Caleb Ziems, Jiaao Chen, Camille Harris, Jessica Anderson, and Diyi Yang. 2022.
\newblock \href {https://doi.org/10.18653/v1/2022.acl-long.258} {{VALUE}: {U}nderstanding dialect disparity in {NLU}}.
\newblock In \emph{Proceedings of the 60th Annual Meeting of the Association for Computational Linguistics (Volume 1: Long Papers)}, pages 3701--3720, Dublin, Ireland. Association for Computational Linguistics.

\bibitem[{Ziems et~al.(2023)Ziems, Held, Yang, Dhamala, Gupta, and Yang}]{ziems2023multivalueframeworkcrossdialectalenglish}
Caleb Ziems, William Held, Jingfeng Yang, Jwala Dhamala, Rahul Gupta, and Diyi Yang. 2023.
\newblock \href {https://arxiv.org/abs/2212.08011} {Multi-value: A framework for cross-dialectal english nlp}.
\newblock \emph{Preprint}, arXiv:2212.08011.

\end{thebibliography}
\newpage
\onecolumn
\appendix

\section{Few-Shot AAVE Translation Examples}
\label{sec:few_shot_examples}

\begin{table*}[ht!]
\centering
\small
\begin{tabular}{P{0.9\textwidth}}
\toprule
\textbf{AAVE Translation Examples} \\
\midrule
"I was bewildered, but I knew dat it was no gud asking his ass to explain." \\
"Cochran pontificated windily for da camera." \\
"I don’t want them to follow in my footsteps, as I ain’t go to no college, but I want them to go." \\
\bottomrule
\end{tabular}
\caption{Examples of the Few-Shot Prompted AAVE Translations from VALUE used in our experiments.}
\label{tab:few_shot_translations_appendix}
\end{table*}


\section{Human Validators' Scores}
\label{sec:human_validators_scores}

\begin{table*}[ht!]
\centering
\small
\setlength{\tabcolsep}{12pt}
\begin{tabular}{cccccc}
\toprule
\textbf{Validator} & \textbf{COPA} & \textbf{BoolQ} & \textbf{MultiRC} & \textbf{SST-2} & \textbf{WSC} \\
\midrule
\textbf{Validator 1}  & 6.9 & 7.1 & 7.4 & 6.6 & 7.3 \\
\textbf{Validator 2}  & 7.0 & 7.4 & 7.5 & 7.4 & 8.3 \\
\textbf{Validator 3}  & 7.6 & 6.5 & 7.0 & 7.2 & 6.9 \\
\textbf{Validator 4}  & 6.5 & 6.9 & 7.0 & 6.6 & 7.0 \\
\textbf{Validator 5}  & 7.5 & 6.9 & 7.4 & 7.0 & 6.9 \\
\textbf{Validator 6}  & 7.3 & 7.0 & 7.2 & 7.3 & 7.2 \\
\textbf{Validator 7}  & 7.6 & 7.2 & 7.5 & 7.5 & 7.1 \\
\textbf{Validator 8}  & 7.2 & 7.1 & 7.3 & 7.1 & 7.4 \\
\textbf{Validator 9}  & 7.3 & 7.2 & 7.4 & 7.3 & 7.3 \\
\textbf{Validator 10} & 7.3 & 6.9 & 7.0 & 6.9 & 7.1 \\
\midrule
\textbf{Average}      & \textbf{7.22} & \textbf{7.02} & \textbf{7.27} & \textbf{7.09} & \textbf{7.25} \\
\bottomrule
\end{tabular}
\caption{Human Validators' Scores for AAVE Translations Across All Tasks (Out of 10)}
\label{tab:human_validators_all_tasks}
\end{table*}


\section{BLEU Scores}
\label{sec:bleu_scores}

\begin{table*}[ht!]
\centering
\small
\setlength{\tabcolsep}{12pt}
\begin{tabular}{cccc}
\toprule
\textbf{Task} & \textbf{BLEU $<$ 0.7 (\%)} & \textbf{BLEU $<$ 0.5 (\%)} & \textbf{BLEU $<$ 0.3 (\%)} \\
\midrule
BoolQ (Passage)       & 91.09\%  & 57.66\%  & 18.82\% \\
BoolQ (Question)      & 79.38\%  & 53.35\%  & 35.64\% \\
COPA (Premise)        & 87.20\%  & 74.80\%  & 59.40\% \\
COPA (Choice 1)       & 85.40\%  & 68.20\%  & 58.60\% \\
COPA (Choice 2)       & 80.60\%  & 68.20\%  & 56.40\% \\
MultiRC (Paragraph)   & 100.00\% & 98.90\%  & 95.00\% \\
MultiRC (Question)    & 77.50\%  & 55.20\%  & 31.10\% \\
SST-2 (Sentence)      & 96.60\%  & 85.70\%  & 64.10\% \\
WSC (Paragraph)       & 88.15\%  & 57.29\%  & 23.71\% \\
\bottomrule
\end{tabular}
\caption{BLEU Scores Across Tasks (Percentage of Translations Below BLEU Thresholds). These scores measure the lexical similarity of our translations to the original texts.}
\label{tab:bleu_scores_appendix}
\end{table*}


\end{document}